\documentclass[runningheads]{llncs}
\usepackage{graphicx}
\usepackage{comment}
\usepackage{amsmath,amssymb} 
\usepackage{color}
\usepackage{caption}
\usepackage[utf8]{inputenc}
\DeclareUnicodeCharacter{00A8}{ }

\usepackage{subfigure}
\usepackage{xcolor,colortbl}
\definecolor{Gray}{gray}{0.85}
\newcolumntype{a}{>{\columncolor{Gray}}c}
\usepackage{floatrow}
\newfloatcommand{capbtabbox}{table}[][\FBwidth]
\usepackage{wrapfig}

\begin{document}
\pagestyle{headings}
\mainmatter
\def\ECCVSubNumber{2201}  

\title{SRNet: Improving Generalization in 3D Human Pose Estimation with a Split-and-Recombine Approach}

\titlerunning{SRNet: Improving Generalization in 3D Human Pose Estimation}

\authorrunning{Zeng et al.}
\author{Ailing Zeng\inst{1,2}\thanks{The work is done when Ailing Zeng is an intern at Microsoft
Research Asia.} \and
Xiao Sun\inst{2}\and
Fuyang Huang\inst{1} \and Minhao Liu\inst{1} \and Qiang Xu\inst{1} \and Stephen Lin\inst{2} }
%
\institute{ The Chinese University of Hong Kong \\
\and
Microsoft Research Asia\\
\email{alzeng, fyhuang, mhliu, qxu@cse.cuhk.edu.hk}\\
\email{xias, stevelin@microsoft.com}}
\maketitle

\begin{abstract}
Human poses that are rare or unseen in a training set are challenging for a network to predict. Similar to the long-tailed distribution problem in visual recognition, the small number of examples for such poses limits the ability of networks to model them. Interestingly, {\em local} pose distributions suffer less from the long-tail problem, i.e., local joint configurations within a rare pose may appear within other poses in the training set, making them less rare. We propose to take advantage of this fact for better generalization to rare and unseen poses. To be specific, our method splits the body into local regions and processes them in separate network branches, utilizing the property that a joint’s position depends mainly on the joints within its local body region. Global coherence is maintained by recombining the global context from the rest of the body into each branch as a low-dimensional vector. With the reduced dimensionality of less relevant body areas, the training set distribution within network branches more closely reflects the statistics of {\em local} poses instead of global body poses, without sacrificing information important for joint inference. The proposed split-and-recombine approach, called {\em SRNet}, can be easily adapted to both single-image and temporal models, and it leads to appreciable improvements in the prediction of rare and unseen poses.
\keywords{Human pose estimation, 2D to 3D, long-tailed distribution}
\end{abstract}

\section{Introduction}
\label{sec:intro}

Human pose estimation is a longstanding computer vision problem with numerous applications, including human-computer interaction, augmented reality, and computer animation. For predicting 3D pose, a common approach is to first estimate the positions of keypoints in the 2D image plane, and then lift these keypoints into 3D. The first step typically leverages the high performance of existing 2D human pose estimation algorithms~\cite{chen2018cascaded,newell2016stacked,sun2019deep}. For the second stage, a variety of techniques have been proposed, based on structural or kinematic body models~\cite{cai2019exploiting,dabral2018learning,fang2018learning,lee2018propagating,wandt2019repnet}, learned dependencies and relationships among body parts~\cite{fang2018learning,park20183d}, and direct regression~\cite{martinez2017simple}.

\begin{figure}[t]
\begin{center}
\includegraphics[width=1.0\textwidth]{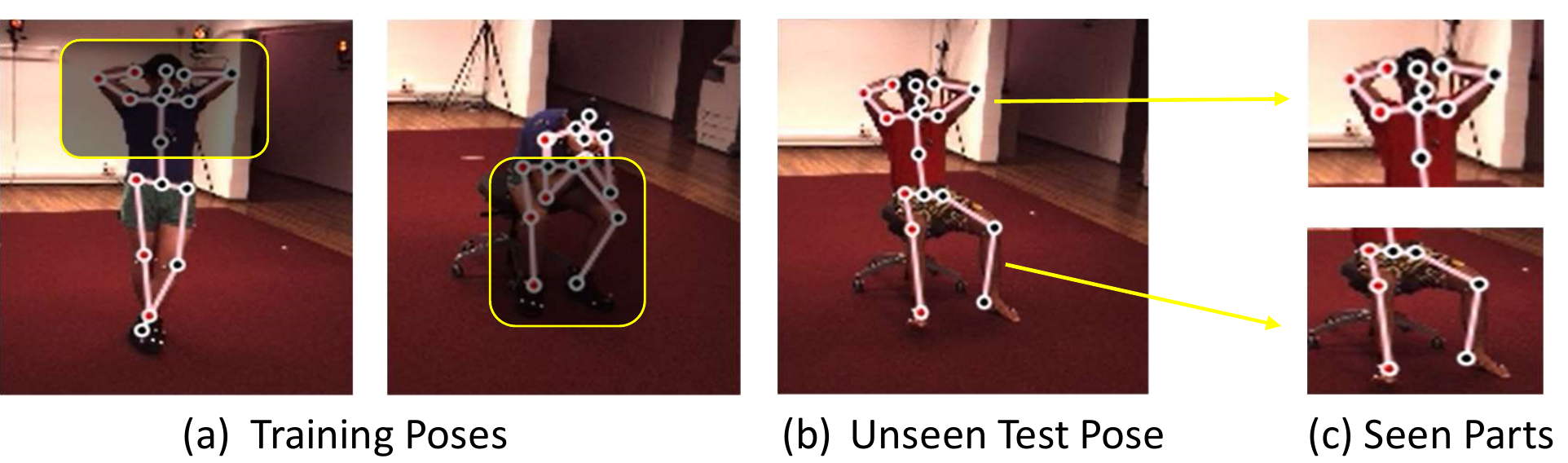}

\end{center}
\caption{An unseen test pose (b) may be decomposed into local joint configurations (c) that appear in poses that exist in the training set (a). Our method takes advantage of this property to improve estimation of rare and unseen poses.}
\label{fig:date_fig}
\end{figure}

Besides algorithm design, an important factor in the performance of a machine learning system is its training data. A well-known issue in pose estimation is the difficulty of predicting poses that are rare or unseen in the training set. Since few examples of such poses are available for training, it is hard for the network to learn a model that can accurately infer them. Better generalization to such poses has been explored by augmenting the training set with synthetically generated images~\cite{rogez2016mocap,chen2016synthesizing,varol2017learning,wang2019generalizing,mehta2017monocular,jahangiri2017generating}. However, the domain gap that exists between real and synthesized images may reduce their efficacy. Different viewpoints of existing training samples have also been simulated to improve generalization to other camera positions~\cite{fang2018learning}, but this provides only a narrow range of pose augmentations.

In this work, we propose to address the rare/unseen pose problem through a novel utilization of the data in the original training set. Our approach is based on the observation that rare poses at the global level are composed of local joint configurations that are generally less rare in the training set. For example, a bicycling pose may be uncommon in the dataset, but the left leg configuration of this pose may resemble the left legs of other poses in the dataset, such as stair climbing and marching. Many instances of a local pose configuration may thus exist in the training data among different global poses, and they could be leveraged for learning local pose. Moreover, it is possible to reconstruct unseen poses as a combination of local joint configurations that are presented in the dataset. For example, an unseen test pose may be predicted from the upper body of one training pose and the lower body of another, as illustrated in Fig.~\ref{fig:date_fig}. 

Based on this observation, we design a network structure that splits the human body into local groups of joints that have strong inter-relationships within each group and relatively weak dependencies on joints outside the group. Each group is processed in a separate network branch. To account for the weak dependencies that exist with the rest of the body, low-dimensional global context is computed from the other branches and recombined into the branch. The dimensionality reduction of less-relevant body areas within the global context decreases their impact on the feature learning for local joints within each branch. At the same time, accounting for some degree of global context can avoid local pose estimates that are incoherent with the rest of the body. With this split-and-recombine approach, called {\em SRNet}, generalization performance is enhanced to effectively predict global poses that are rare or absent from the training set. 


In extensive comparisons to state-of-the-art techniques, SRNet exhibits competitive performance on single-frame input and surpasses them on video input. More importantly, we show that SRNet can elevate performance considerably on rare/unseen poses. Moreover, we conduct various ablation studies to validate our approach and to examine the impact of different design choices.

\section{Related Work}
\label{sec:related_work}

Extensive research has been conducted on reconstructing 3D human pose from 2D joint predictions. In the following, we briefly review methods that are closely related to our approach. 


\paragraph{\textbf{Leveraging Local Joint Relations}}

Many recent works take advantage of the physical connections that exist between body joints to improve feature learning and joint prediction. As these connections create strong inter-dependencies and spatial correlations between joints, they naturally serve as paths for information sharing~\cite{lee2018propagating,zhao2019semantic,cai2019exploiting,ci2019optimizing} and for encoding kinematic~\cite{fang2018learning} and anatomical~\cite{dabral2018learning,habibie2019wild,pavllo20193d,wandt2019repnet} constraints.

The structure of these connections defines a locality relationship among joints. More closely connected joints have greater inter-dependency, while distantly connected joints have less and indirect dependence via the joints that lie on the paths between them. These joint relationships have been modeled hierarchically, with feature learning that starts within local joint groups and then expands to account for all the joint groups together at a global level~\cite{park20183d}. Alternatively, these relationships have been represented in a graph structure, with feature learning and pose estimation conducted in a graph convolutional network (GCN)~\cite{zhao2019semantic,cai2019exploiting,ci2019optimizing}. Within a GCN layer, dependencies are explicitly modeled between connected joints and expand in scope to more distant joints through the stacking of layers.

For both hierarchical and GCN based techniques, feature learning within a local group of joints can be heavily influenced by joints outside the group. In contrast, the proposed approach SRNet restricts the impact of non-local joints on local feature learning through dimensionality reduction of the global context, which facilitates the learning of local joint configurations without ignoring global pose coherence. By doing so, SRNet can achieve better generalization to rare and unseen poses in the training data.

\paragraph{\textbf{Generalization in Pose Estimation}}

A common approach to improve generalization in pose estimation is to generate more training images through data augmentation. Approaches to augmentation have included computer graphics rendering~\cite{chen2016synthesizing,varol2017learning}, image-based synthesis~\cite{rogez2016mocap}, fitting 3D models to people and deforming them to generate new images~\cite{pishchulin2012articulated}, and changing the background, clothing and occluders in existing images~\cite{mehta2017monocular}. These methods can produce a large amount of data for training, but the gap in realism between artificially constructed images and actual photographs may limit their effectiveness. Since our technique obtains improved training data distributions by focusing on local pose regions instead of global poses, it does not involve image manipulation or synthesis, thereby maintaining the realism of the original training set.

Other works have explored generalization to different viewpoints~\cite{fang2018learning,veges20193d} and to in-the-wild scenes~\cite{yang20183d,habibie2019wild}, which are orthogonal to our goal of robustly estimating rare and unseen poses.

\paragraph{\textbf{Robustness to Long-tailed Distributions}}

In visual recognition, there exists a related problem of long-tailed training set distributions, where classes in the distribution tail have few examples. As these few examples have little impact on feature learning, the recognition performance for tail classes is often poor. Recent approaches to this problem include metric learning to enforce inter-class margins~\cite{huang2016learning} and meta-learning that learns to regress many-shot model parameters from few-shot model parameters~\cite{wang2017learning}. Such techniques improve the discriminability of tail classes from other classes, but they are not compatible with the problem of keypoint localization in pose estimation.

\begin{figure}[b]
\begin{center}
\includegraphics[width=1\textwidth]{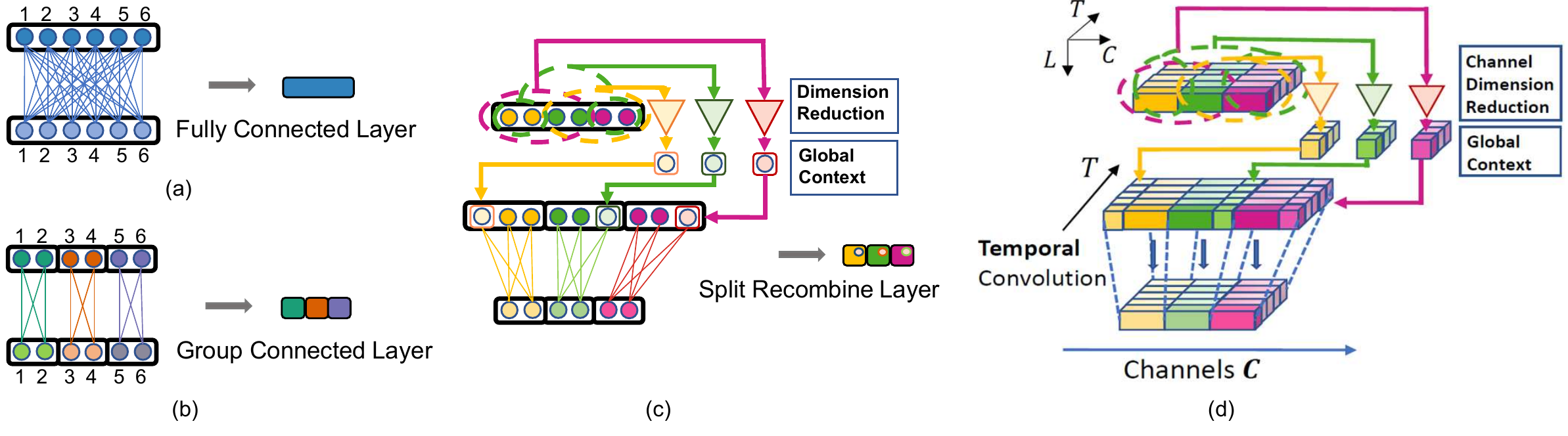}
\end{center}
  \caption{Illustration of (a) a fully connected layer, (b) group connected layer, (c) our split-and-recombine layer, and (d) our convolution layer for temporal models. The four types of layers can be stacked to form different network structures as shown in Figure \ref{fig:network}. }
\label{fig:layer}
\end{figure}

\section{Method}
\label{sec:method}
To address the issue of rare and unseen poses, our approach is to decompose global pose estimation into a set of local pose estimation problems. For this strategy to be effective, the local problems must be defined in a manner that allows the feature learning of a local region to primarily reflect the statistical distribution of its local poses, yet account for other local regions such that the final overall pose estimate is globally coherent.

In the following, we start with the most common baseline for lifting 2D joints to 3D, and then describe how to modify this baseline to follow our strategy.



\paragraph{\textbf{Fully-Connected Network Baseline}}

For lifting 2D keypoints to 3D, a popular yet simple baseline is to use a fully-connected network (FCN) consisting of several layers~\cite{martinez2017simple}. Given the 2D keypoint detections $\mathcal{K} = \{K_i|i,...,N\} \in \mathbb{R}^{2N}$ in the image coordinate system, the FCN estimates the 3D joint locations $\mathcal{J} = \{J_i|i,...,N\} \in \mathbb{R}^{3N}$ in the camera coordinate system with the origin at the root joint $J_0$. Formally, a fully-connected network layer can be expressed as 
\begin{equation}
    \mathbf{f}^{l+1} = \Theta^{l} \mathbf{f}^{l}
\label{eq:fc_layer}
\end{equation}
where $\Theta^{l} \in \mathbb{R}^{D^{l+1} \times D^{l}}$ is the fully-connected weight matrix to be learned, and $D^l$ is the feature dimension for the $l^{th}$ layer, namely $\mathbf{f}^{l} \in \mathbb{R}^{D^l}$. Batch normalization and ReLU activation are omitted for brevity.
For the input and output layers, their feature dimensions are $D^1=2N$ and $D^{L+1}=3N$, respectively. 

It can be noted that in this FCN baseline, each output joint and each intermediate feature is connected to all of the input joints indiscriminately, allowing the prediction of an output joint to be overfitted to the positions of distant joints with little relevance. In addition, all the output joints share the same set of features entering the final layer and have only this single linear layer ($\Theta^{L}_i$) to determine a solution particular to each joint.

\paragraph{\textbf{Body Partitioning into Local Pose Regions}}
\label{sec:modify}
In turning global pose estimation into several local pose estimation problems, a suitable partitioning of the human body into local pose regions is needed. A local pose region should contain joints whose positions are heavily dependent on one another, but less so on joints outside the local region.

Here, we adopt the partitioning used in~\cite{park20183d}, where the body is divided into left/right arms, left/right legs, and torso. These parts have distinctive and coordinated behaviors such that the joint positions within each group are highly correlated. In contrast, joint positions between groups are significantly less related.

To accommodate this partitioning, the FCN layers are divided into groups. Formally, $\mathcal{G}^l_g$ represents the feature/joint indexes of the $g^{th}$ group at layer $l$. Specifically, for the first input layer,
\begin{equation}
    \mathcal{G}^1_0 = [\text{joint indices of the right arm}],
\end{equation}
and for the intermediate feature layers, we have
\begin{equation}
    \mathcal{G}^l_0 = [\text{feature indices of the right leg at the $l_{th}$ layer}].
\end{equation} Then, a \emph{group connected layer} for the $g_{th}$ group is expressed as
\begin{equation}
    \mathbf{f}^{l+1}[\mathcal{G}_g^{l+1}]=\Theta^{l}_g\mathbf{f}^{l}[\mathcal{G}_g^l].
\label{eq:group_layer}
\end{equation}

In a \emph{group connected layer}, the connections between joint groups are removed. This ``local connectivity” structure is commonly used in convolutional neural networks to capture spatially local patterns~\cite{krizhevsky2012imagenet}. The features learned in a group depend on the statistical distribution of the training data for only its local pose region. In other words, this local feature is learned independently of the pose configurations of other parts, and thus it generalizes well to any global pose that includes similar local joint configurations.

However, a drawback of group connected layers is that the status of the other body parts is completely unknown when inferring the local pose. As a result, the set of local inferences may not be globally coherent, leading to low performance. There is thus a need to account for global information while largely preserving local feature independence.

\paragraph{\textbf{SRNet: Incorporating Low-Dimensional Global Context}}
To address this problem, we propose to incorporate \emph{Low-Dimensional Global Context} (LDGC) in a group connected layer. It coarsely represents information from the less relevant joints, and is brought back to the local group in a manner that limits disruption to the local pose modeling while allowing the local group to account for non-local dependencies. This split-and-recombine approach for global context can be expressed as the following modification of Eq.~\ref{eq:group_layer}:
\begin{equation}
    \mathbf{f}^{l+1}[\mathcal{G}_g^{l+1}]=\Theta^{l}_g (\mathbf{f}^{l}[\mathcal{G}_g^l] \circ \mathcal{M}\mathbf{f}^{l}[\mathcal{G}^l \setminus \mathcal{G}_g^l])
\end{equation}    
where $\mathbf{f}^{l}[\mathcal{G}^l \setminus \mathcal{G}_g^l]$ is the global context for the $g^{th}$ group. $\mathcal{M}$ is a mapping function that defines how the global context is represented. Special cases of the mapping function are $\mathcal{M} = Identity$, equivalent to a fully-connected layer, and $\mathcal{M} = Zero$, which is the case for a group connected layer. The mapped global context is recombined with local features by an operator $\circ$, typically concatenation for an \emph{FCN}.

The mapping function $\mathcal{M}$ acts as a gate controlling the information passed from the non-local joints. If the gate is wide open (FCN), the local feature learning will account for the exact joint positions of other body parts, weakening the ability to model the local joint configurations of rare global poses. However, if the gate is closed (Group), the local feature learning receives no knowledge about the state of other body parts and may lose global coherence. We argue that the key to achieving high performance and strong generalization to rare poses is to learn a low-dimensional representation of the global context, namely
\begin{equation}
\mathcal{M} = \Gamma_g^l \in \mathbb{R}^{H \times (D^l - D^l_g)}
\end{equation}
where $D^l_g$ is the feature dimension for the $g$th group at layer $l$. $\Gamma_g^l$ is a weight matrix that maps the global context $\mathbf{f}^{l}[\mathcal{G}^l \setminus \mathcal{G}_g^l]$ of dimensions $D^l - D^l_g$ to a small number of dimensions $H$.

In our experiments, we empirically evaluate different design choices for the mapping function $\mathcal{M}$ and the combination operator $\circ$.



\begin{figure}[t]
\begin{center}
\includegraphics[width=0.8\textwidth]{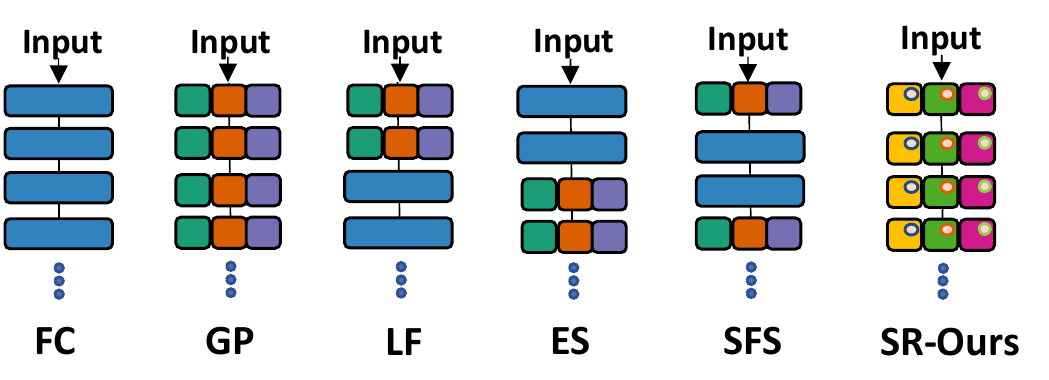}
\end{center}
  \caption{Illustration of fully connected (FC), group connected (GP), late fusion (LF), early split (ES), split-fuse-split (SPS) and split-and-recombine (SR) models. The components of each layer are shown in Figure \ref{fig:layer}.}
\label{fig:network}
\end{figure}

\paragraph{\textbf{Network Structure}}
With our split-and-recombine approach for processing local pose regions and global context, the \emph{SRNet} model can be illustrated as shown at the right side of Fig.~\ref{fig:network}. This network follows the structure of a group connected network, with the body joints split into separate groups according to their local pose region. Added to this network are connections to each group in a layer from the other groups in the preceding layer. As illustrated in Fig.~\ref{fig:layer} (c,d), these connections representing the global context reduce the dimensionality of the context features and recombine them with the local features in each group. These inputs are mapped into outputs of the original feature dimensions.

\emph{SRNet} balances feature learning dependency between the most related local region and the less related global context. For demonstrating the efficacy of this split-and-recombine strategy, we describe several straightforward modifications to the FC network, illustrated in Fig.~\ref{fig:network}, that will serve as baselines in our experiments:
\begin{itemize}
\item \emph{Late Fusion (LF)}, which postpones feature sharing among groups by cutting their connections in the earlier layers. This structure is similar to that used in~\cite{park20183d}, which first learns features within each group and fuses them later with stacked fully connected layers. Formally, the \emph{LF} baseline is defined as
\begin{equation}
    \mathbf{f}^{l+1}= 
\begin{cases}
    Cat\{\mathbf{f}^{l}[\mathcal{G}_g^{l}]|g=1,...,G\},& \text{if } l < L_{fuse}\\
    \Theta^{l}\mathbf{f}^{l},              & \text{otherwise.}
\end{cases}
\end{equation}

\item \emph{Early Split (ES)}, which aims to provide more expressive features for each group, by cutting the connections between groups in the latter layers. Formally,
\begin{equation}
    \mathbf{f}^{l+1}= 
\begin{cases}
    \Theta^{l}\mathbf{f}^{l},& \text{if } l < L_{split}\\
    Cat\{\mathbf{f}^{l}[\mathcal{G}_g^{l}]|g=1,...,G\},              & \text{otherwise.}
\end{cases}
\end{equation}

\item \emph{Group Connected (GP)}, which is the standard group connected network. \emph{Late Fusion} degenerates to this form when $L_{fuse}=L$, and \emph{Early Split} does when $L_{split}=1$.

\item \emph{Split-Fuse-Split (SFS)}, where the middle $L_{link}$ layers of the network are fully-connected and the connections between joint groups are cut in the first and last layers.
\end{itemize}

The differences among FCN, these baselines, and SRNet are summarized in Table~\ref{tab:Summary}. The table also highlights differences of SRNet from GCN~\cite{zhao2019semantic} and its modification LCN~\cite{ci2019optimizing}, specifically in controlling the feature dependencies between local regions and global context.

\begin{table}
{
\scriptsize{
\begin{center}
\begin{tabular}{ l |c|c|c|c}
\hline
Method      & local inference & local features & info passing & dimension control\\
\hline
FCN         & weak(share feature) & no  & yes & no  \\
Group       & yes & yes & no  & no  \\
\hline
Late Fusion~\cite{park20183d} & weak(share feature) & yes & yes & no  \\
Early Split & yes & no  & yes & no  \\
Split-Fuse-Split & yes & yes & yes & no  \\
\hline
GCN~\cite{zhao2019semantic}  & no(share weight)  & yes(overlapped group) & yes & no  \\
LCN~\cite{ci2019optimizing}  & yes & yes(overlapped group)  & yes & no  \\
\hline
Ours (SRNet) & yes & yes & yes & \textbf{yes} \\
\hline
\end{tabular}
\end{center}
}
}
\caption{Different network structures used for 2D to 3D pose estimation.}
\label{tab:Summary}
\end{table}

\paragraph{\textbf{SR Convolution for Temporal Models}} Temporal information in 2D to 3D mapping is conventionally modeled with Recurrent Neural Networks (RNNs)~\cite{cai2019exploiting,lee2018propagating}. Recently, Pavllo et al.~\cite{pavllo20193d} introduce an alternative temporal convolutional model that stacks all the spatial information of a frame into the channel dimensions and replaces fully-connected operations in the spatial domain by convolutions in the temporal domain. The temporal convolution enables parallel processing of multiple frames and brings greater accuracy, efficiency, and simplicity.

As illustrated in Fig.~\ref{fig:layer} (d), our split-and-recombine modification to the fully connected layer can be easily adapted to the temporal convolution model by applying the same split-and-recombine strategy to the channels during convolution. 
Specifically, the group connected operations for local joint groups are replaced with corresponding temporal convolutions for local feature learning, where the features from other joint groups undergo an extra convolution for dimension reduction in channels and are concatenated back with the local joint group as global context.
We call this split-and-recombine strategy in the channel dimensions during convolution the \emph{SR convolution}.

\section{Datasets and Rare-Pose Evaluation Protocols}
\label{sec:data}
\subsection{Datasets and Evaluation Metrics}
Our approach is validated on two popular benchmark datasets:

\noindent -- \emph{Human3.6M} \cite{h36m_pami} is a large benchmark widely-used for 3D human pose estimation. It consists of 3.6 million video frames from four camera viewpoints with 15 activities. Accurate 3D human joint locations are obtained from motion capture devices. Following convention, we use the mean per joint position error (MPJPE) for evaluation, as well as the Procrustes Analysis MPJPE (PA-MPJPE).

\noindent -- \emph{MPI-INF-3DHP} \cite{mehta2017monocular,mehta2017vnect} is a more challenging 3D pose dataset, containing not only constrained indoor scenes but also complex outdoor scenes. Compared to Human3.6M, it covers a greater diversity of poses and actions. 
For evaluation, we follow common practice~\cite{wandt2019repnet,ci2019optimizing,wang2019generalizing,yang20183d} by using the Percentage of Correct Keypoints (PCK) with a threshold of 150mm and the Area Under Curve (AUC) for a range of PCK thresholds.




\subsection{Evaluation Protocols}
\label{eval_set}
The most widely-used evaluation protocol~\cite{martinez2017simple,wang2019generalizing,huang2019deepfuse,pavllo20193d,ci2019optimizing} on Human3.6M uses five subjects (S1, S5, S6, S7, S8) as training data and the other two subjects (S9, S11) as testing data. We denote it as the \emph{\textbf{Subject Protocol}}. However, the rare pose problem is not well examined in this protocol since each subject is asked to perform a fixed series of actions in a similar way. 


To better demonstrate the generalization of our method to rare/unseen poses, we use two other protocols, introduced in~\cite{ci2019optimizing}. The first is the \emph{\textbf{Cross Action Protocol}} that trains on only one of the 15 actions in the Human3.6M dataset and tests on all actions. The second is the \emph{\textbf{Cross Dataset Protocol}} that applies the model trained on Human3.6M to the test set of MPI-INF-3DHP. In addition, we propose a new protocol called \emph{\textbf{Rare Pose Protocol}}. 
\paragraph{\textbf{Rare Pose Protocol}.} In this protocol, we use all the poses of subjects S1, S5, S6, S7, S8 for training but only use a subset of poses from subjects S9, S11 for testing. The subset is selected as the rarest poses in the testing set.
To identify rare poses, we first define pose similarity ($PS$) as
\begin{equation}
    PS(\mathcal{J}, \mathcal{I}) = \frac{1}{N}\sum_{i}^{N} exp(-\dfrac{||J_i - I_i||^2}{2\sigma_i^2} )
\end{equation}
where $||J_i - I_i||$ is the Euclidean distance between corresponding joints of two poses. To compute $PS$, we pass $\Delta J_i$ through an unnormalized Gaussian with a standard deviation $\sigma_i$ controlling falloff. This yields a pose similarity that ranges between 0 and 1. Perfectly matched poses will have $PS=1$, and if all joints are distant by more than the standard deviation $\sigma_i$, $PS$ will be close to 0. The occurrence of a pose $\mathcal{J}$ in a pose set $\mathbb{O}$ is defined as the average similarity between itself and all the poses in $\mathbb{O}$:
\begin{equation}
    OCC_{\mathbb{O}}(\mathcal{J}) = \dfrac{1}{M} \sum_{\mathcal{I}}^{\mathbb{O}} PS(\mathcal{J}, \mathcal{I}) 
\end{equation}
where $M$ is the total number of poses in the pose set $\mathbb{O}$. A rare pose will have a low occurrence value with respect to other poses in the set. We select the $R\%$ of poses with the lowest occurrence values in the testing set for evaluation. 






\section{Experiments}
\label{sec:result}

\subsection{Ablation Study}
\label{sec:ablation}
All our ablation studies are evaluated on the Human3.6M dataset. Both the \emph{Subject Protocol} and our new \emph{Rare Pose Protocol} are used for evaluation. MPJPE serves as the evaluation metric. \emph{Please refer to the supplementary material for details on data pre-processing and training settings.}


\begin{table}[b]
{
\begin{center}
\begin{tabular}{ l |a| c|| c |c |c ||c }
\hline
 Protocol & GP& FC & LF & ES & SFS  & Ours (SR)\\
\hline
Subject (100\%) & 62.7 & \underline{46.8} &39.8 &42.0&$\underline{39.4}\downarrow_{7.4}$&$\textbf{36.6}\downarrow_{2.8(7.1\%)}$\\
\hline
Rare Pose (20\%)& 89.1 & \underline{76.0} &  60.1&62.5 & $\underline{59.2}\downarrow_{16.8}$ &$\textbf{  48.6}\downarrow_{10.6(17.9\%)}$\\
\hline
Rare Pose (10\%)& 98.9& \underline{89.1} &69.9 &73.5 &$\underline{68.2}\downarrow_{20.9}$ &$\textbf{  53.7}\downarrow_{14.5(21.3\%)}$\\
\hline
\end{tabular}
\end{center}
}
\caption{Comparing the \emph{SR} network to different baseline networks under the \emph{Subject} protocol and the \emph{Rare Pose} protocol (with 10\% and 20\% of the rarest poses). MPJPE is used as the evaluation metric. The improvements of \emph{SFS} from \emph{FC}, and of \emph{SR} from \emph{SFS}, are shown as subscripts.}
\label{tab:srpose}
\end{table}

\paragraph{\textbf{Effect of the SRNet}} In Table~\ref{tab:srpose}, we compare our \emph{SR} network with the baseline networks \emph{FC}, \emph{GP}, \emph{LF}, \emph{ES}, \emph{SFS} described in Section~\ref{sec:method}. By default, the mapped dimensions $H$ is set to 1 and the combination operator $\circ$ is set to multiplication.

\begin{figure}[t]
\includegraphics[width=1.0\linewidth]{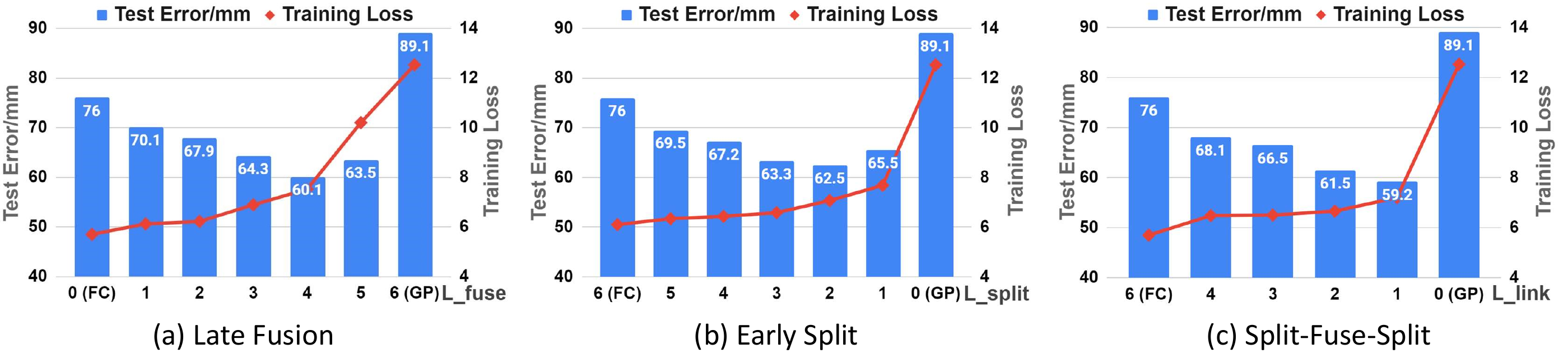}
\caption{Training loss and test error (MPJPE in mm) of the baseline networks (a) \emph{LF}, (b) \emph{ES}, and (c) \emph{SFS} with respect to different splitting configurations. Rare Pose Protocol 20\% is used for evaluation. }
\label{fig:baseline_performance}
\end{figure}

It is found that both the \emph{LF} and \emph{ES} baselines are superior to the \emph{FC} and \emph{GP} baselines. A more detailed comparison is shown in Figure~\ref{fig:baseline_performance}, with different splitting configurations for each. In Figure~\ref{fig:baseline_performance}, both \emph{LF} and \emph{ES} start on the left with the \emph{FC} network ($L_{fuse} = 0$ and $L_{split} = 6$). When the fully connected layers are gradually replaced by group connected layers, from the beginning of the network for $LF$ ($L_{fuse}\uparrow$) or the end for $ES$ ($L_{split}\downarrow$), testing error decreases at first. This indicates that \emph{local feature learning} and \emph{local inference} are each helpful. However, replacing all the fully connected layers with group layers leads to a sharp performance drop. This indicates that a local joint region still needs information from the other regions to be effective.

Combining the \emph{LF} and \emph{ES} baselines to create \emph{SFS} yields further improvements from sharing the merits of both. These two simple modifications already improve upon \emph{FC} by a large margin. Specifically, it improves upon \emph{FC} by 7.4mm (relative 15.8\%), 16.8mm (relative 22.1\%) and 20.9mm (relative 23.5\%) on the \emph{Subject}, \emph{Rare Pose 20\%} and \emph{Rare Pose 10\%} protocols, respectively. 

\begin{figure*}[tb]
\begin{center}
\includegraphics[width=1\textwidth]{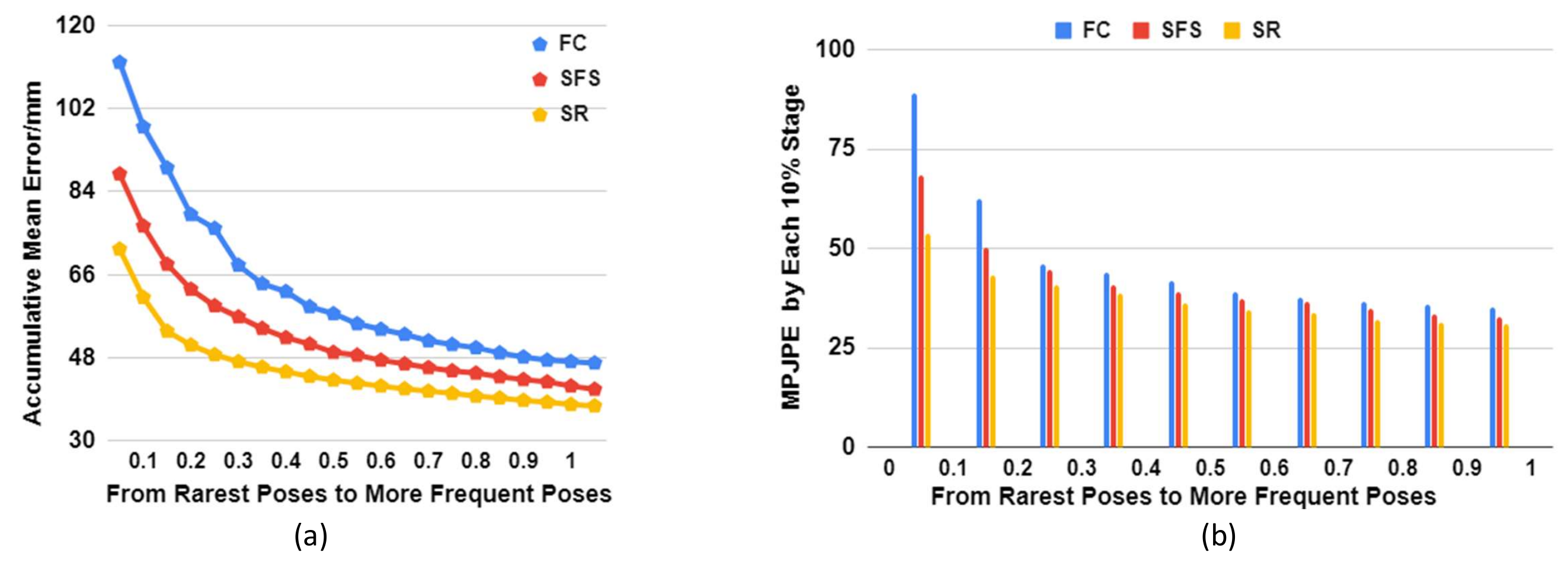}
\end{center}
\caption{Testing errors of \emph{FC}, \emph{SFS} and our \emph{SR} networks on poses with different degrees of rareness. The horizontal axis represents the top $R\%$ of rarest poses. (a) Cumulative mean error. (b) Mean error at $10\%$ intervals.}
\label{fig:nn_result}
\end{figure*}

\emph{SRNet} with its use of low-dimensional global context performs significantly better than the \emph{SFS} baseline. Specifically, it improves upon \emph{SFS} by 2.8mm (relative 7.1\%), 10.6mm (relative 17.9\%) and 14.5mm (relative 21.2\%) on the \emph{Subject}, \emph{Rare Pose 20\%} and \emph{Rare Pose 10\%} protocols, respectively. Note that the improvement is especially large on rare poses, indicating the stronger generalization ability of $SRNet$ to such poses. Figure~\ref{fig:nn_result} breaks down the results of the \emph{FC}, \emph{SFS} and \emph{SR} networks for different degrees of pose rareness. It can be seen that the performance improvements of SRNet increase for rarer poses.

\begin{table}[h]
\scriptsize{
\begin{center}
\begin{tabular}{ l |c |c|c|c|c|c|c|c}
\hline
Dimension $H$ &0(GP)& 1 & 3 & 10 & 25\% $D^{l}_g$ &50\% $D^{l}_g$&100\% $D^{l}_g$&200\% $D^{l}_g$\\
\hline
Subject (100\%) &62.7&\textbf{38.3} &41.9&42.9&44.6&45.3&45.7&45.8\\
\hline
Rare Pose (20\%) &89.1&\textbf{49.1}&52.2&53.6&56.5&63.8&70.1&78.5\\
\hline
\end{tabular}
\end{center}
}
\caption{Effect of using different dimension $H$ for the proposed global context. $100\%$ corresponds to the same dimensionality as the local feature.}
\label{tab:low_dim}
\end{table}

\paragraph{\textbf{Low-Dimensional Representation for Global Context}}
Table~\ref{tab:low_dim} presents the performance of SRNet with different dimensions $H$ for the global context. To readily accommodate different dimensions, the combination operator $\circ$ is set to concatenation. When $H=0$, \emph{SRNet} degenerates into the \emph{GP} baseline. It can be seen that the \emph{GP} baseline without any global context obtains the worst performance. This shows the importance of bringing global context back to local groups. Also, it is found that a lower dimension of global context leads to better results. The best result is achieved by keeping \emph{only a single dimension} for global context. This indicates that a low-dimensional representation of the global context can better account for the information of less relevant joints while preserving effective local feature learning.


\begin{figure}
\scriptsize
\begin{floatrow}
\capbtabbox[.55\textwidth]
{
    \begin{tabular}{ l |c| c| c |c |c|c|c }
    \hline
    Group Num &1(FC)& 2&  3  &5&6&8&17\\
    \hline
    Subj.(100\%) &46.8&41.4 & 37.7& \textbf{36.6}&41.1&46.7&91.5\\
    \hline
    Rare(20\%)&76.0&55.9 &51.0 &\textbf{48.6}&55.6&60.6&115.3\\
    \hline
    \end{tabular}
%
}{%
\caption{Mean testing error with different group numbers for \emph{Subject} and \emph{Rare Pose $20\%$} protocols.}%
\label{tab:group_num}
}
\capbtabbox[.45\textwidth]
{%
    \begin{tabular}{  l |c| c| c |c|c}
    \hline
    Shuffled Groups &5 &4&3&2& 0(Ours)\\
    \hline
    Subject (100\%) &53.8 & 49.7&45.9 &43.4 &\textbf{36.6}\\
    \hline
    \end{tabular}
}{%
\caption{Mean testing error with respect to the number of shuffled groups (5 in total).}
\label{tab:shuffle_tab}
}

\end{floatrow}
\end{figure}

\begin{figure}
\begin{center}
\includegraphics[width=1\textwidth]{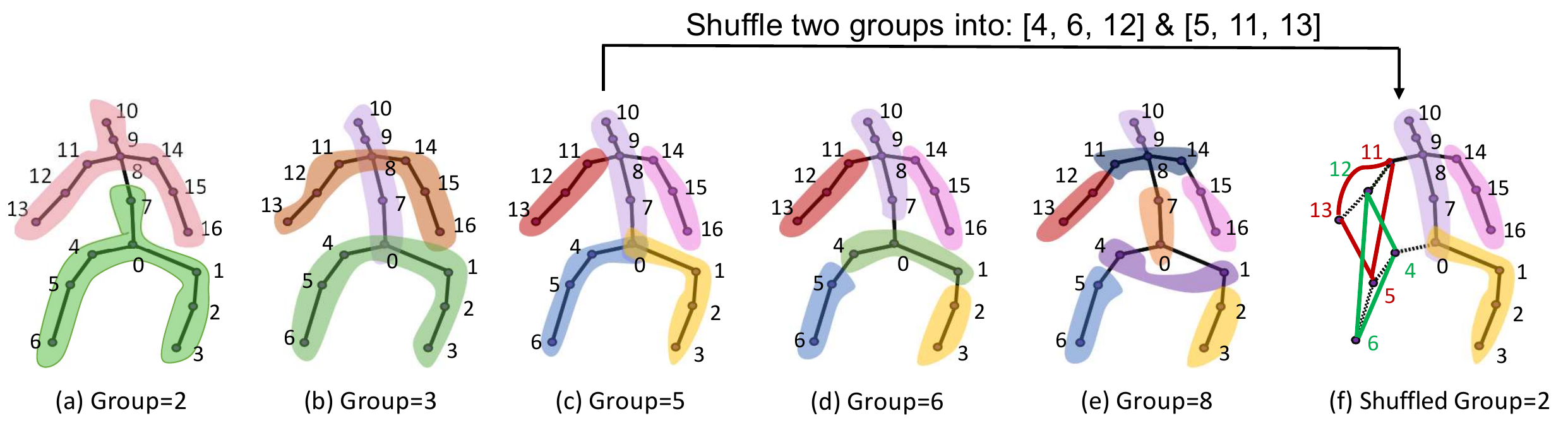}
\end{center}
\caption{Different numbers of groups and their corresponding sets of joints. Joints in the same group are shown in the same color. A random shuffle setting is also shown and evaluated in Table~\ref{tab:shuffle_tab}.}
\label{fig:group_fig}
\end{figure}

\paragraph{\textbf{Effect of Grouping Strategy}} We compare the results of using different numbers of local groups in Table~\ref{tab:group_num}. The corresponding sets of joints for each group are shown in Figure~\ref{fig:group_fig}. More joint groups leads to greater local correlation of joints within a group. It is shown that the performance first improves with more groups, especially in the \emph{Rare Pose} case. However, with more than five groups,
the performance drops due to weakening of local features when there are fewer than three joints in a group.
Moreover, to show that separation into {\em local} groups is important, we randomly shuffle the joints among a subset of groups, with five groups in total, as illustrated in Figure~\ref{fig:group_fig} (f). In Table~\ref{tab:shuffle_tab}, it is shown that the performance decreases when randomly shuffling joints among more groups. This indicates that a strong physical relationship among joints in a group is a prerequisite for learning effective local features.

\subsection{Comparison with State-of-The-Art Methods}
\label{sec:data_result}
We compare with the state of the art in three different experiment settings: cross-dataset on MPI-INF-3DHP, single-frame model on Human3.6M, and temporal model on Human3.6M. MPJPE is used as the evaluation metric on Human3.6M. \emph{Please refer to the supplementary material for more results that use PA-MPJPE as the evaluation metric.}

\noindent\textbf{Cross-Dataset Results on MPI-INF-3DHP.} In this setting, we follow common practice~\cite{luo2018orinet,wang2019generalizing,ci2019optimizing,yang20183d,zhou2017towards} by training the model on the Human3.6M training set and evaluating on the MPI-INF-3DHP test set to examine cross-dataset generalization ability. PCK and AUC are used as evaluation metrics, for which higher is better. In Table \ref{tab:mpi_result}, our approach achieves the best cross-data generalization performance. It improves upon the state-of-the-art~\cite{ci2019optimizing} by a large margin (4.9\% on PCK and 19.3\% on AUC), indicating superior generalization ability. 
\begin{table*}
{
\scriptsize{
\begin{center}
\resizebox{\textwidth}{7mm}
{
\begin{tabular}{ l |c| c|c|c| c |c|c|c|c}
\hline
Method &Martinez \cite{martinez2017simple}&Mehta \cite{mehta2017monocular}&Luo \cite{luo2018orinet}&Biswas \cite{biswas2019lifting}&Yang \cite{yang20183d}&Zhou \cite{zhou2017towards}&Wang \cite{wang2019generalizing}&Ci \cite{ci2019optimizing}&Ours\\
\hline
Outdoor &31.2&58.8&65.7&67.4&-&72.7&-&77.3&$\textbf{80.3}\uparrow_{3.9\%}$\\
\hline
PCK &42.5&64.7&65.6&65.8&69.0&69.2&71.2&74.0& $\textbf{77.6}\uparrow_{4.9\%}$\\
\hline
AUC& 17.0&31.7&33.2&31.2&32.0&32.5&33.8&36.7&$\textbf{43.8}\uparrow_{19.3\%}$\\
\hline
\end{tabular}}
\end{center}
}
}
\caption{Cross-dataset results on MPI-INF-3DHP. All models are trained on Human3.6M and tested on the MPI-INF-3DHP test set.}
\label{tab:mpi_result}
\end{table*}

\begin{table*}
\begin{center}
\resizebox{\textwidth}{7mm}
{
\begin{tabular}{ l c c c c c c c c c c c c c c c c }
\hline
Method & Direct & Discuss & Eat & Greet & Phone& Photo & Pose  & Purcha. & Sit & SitD &Smoke &Wait &WalkD&Walk& WalkT & Avg.\\
\hline
Martinez \cite{martinez2017simple}&127.3&104.6&95.0&116.1&95.5&117.4&111.4&125.0&116.9&93.6&111.0&126.0&131.0&106.3&140.5&114.5\\
\hline
Ci \cite{ci2019optimizing}&102.8&88.6&79.1&99.3&80.0&91.5&93.2&89.6&90.4&\textbf{76.6}&89.6&102.1&108.8&90.8&118.9&93.4\\
\hline
Ours &\textbf{92.3}&\textbf{71.4}&\textbf{71.8}&\textbf{86.4}&\textbf{66.8}&\textbf{79.1}&\textbf{82.5}&\textbf{86.6}&\textbf{88.9}&93.4&\textbf{66.1}&\textbf{83.0}&\textbf{74.4}&\textbf{90.0}&\textbf{97.8}&\textbf{82.0}\\
\hline
\end{tabular}}
\end{center}
\caption{\emph{Cross Action} results compared with \emph{Fully Connected Network} and \emph{Locally Connected Network} on Human3.6M. Smaller values are better. }
\label{tab:lcn}
\end{table*}

\begin{table*}
\begin{center}
\resizebox{\textwidth}{4mm}
{
\begin{tabular}{ l |c| c| c| c |c |c |c |c |c}
\hline
Method &Luvizon\cite{luvizon20182d}&Martinez\cite{martinez2017simple}&Park\cite{park20183d}&Wang \cite{wang2019generalizing}&Zhao\cite{zhao2019semantic}&Ci\cite{ci2019optimizing}&Pavllo \cite{pavllo20193d} &Cai\cite{cai2019exploiting}&Ours\\
\hline
Subject (100\%) &64.1&62.9&58.6&58.0&57.6&52.7&51.8&50.6&\textbf{49.9}\\
\hline
\end{tabular}}
\end{center}
\caption{Comparison on \emph{single-frame 2D pose detection input} in terms of mean per-joint position error (MPJPE). Best result in bold. }
\label{tab:human_p1_sd}
\end{table*}

\begin{table*}
\begin{center}
\scriptsize
{
\begin{tabular}{ l| c |c |c |c |c |c|c }
\hline
Method &Martinez \cite{martinez2017simple}&Pham \cite{pham2019unified}&Biswas\cite{biswas2019lifting} &Zhao \cite{zhao2019semantic}&Wang \cite{wang2019generalizing} &Ci \cite{ci2019optimizing} &Ours \\
\hline
Subject (100\%) &45.5&42.4&42.8&43.8&37.6&36.4&\textbf{33.9}\\
\hline
\end{tabular}}
\end{center}
\caption{Comparison on \emph{single-frame 2D ground truth pose input} in terms of MPJPE. With ground truth 2D pose as input, the upper bound of these methods is explored. 
}
\label{tab:human_p1_ss}
\end{table*}

\noindent\textbf{Single-Frame Model Results on Human3.6M.} In this setting, 
we first compare with the state-of-the-art method~\cite{ci2019optimizing} using the \emph{Cross Action Protocol}. Table \ref{tab:lcn} shows that our method yields an overall improvement of 11.4mm (relative 12.2\% improvement) over~\cite{ci2019optimizing} and performs better on 93\% of the actions, indicating strong generalization ability on unseen actions. We also compare our model to previous works using the standard \emph{Subject Protocol} in Table~\ref{tab:human_p1_sd} and Table~\ref{tab:human_p1_ss}, which use 2D keypoint detection and 2D ground truth as inputs, respectively. 
Under the standard protocol, our model surpasses the state-of-the-art in each case, namely~\cite{cai2019exploiting} for 2D keypoint detection and~\cite{ci2019optimizing} for 2D ground truth as input.

\noindent\textbf{Temporal Model Results on Human3.6M} In Table~\ref{tab:human_p1_t}, our approach achieves the new state-of-the-art with either 2D keypoint detection or 2D ground truth (with $\bigtriangledown$) as input. Specifically, SRNet improves upon \cite{pavllo20193d} from 37.2mm to 32.0mm (relative 14.0\% improvement) with 2D ground truth input and from 46.8mm to 44.8mm (relative 4.3\% improvement) with 2D keypoint detection input. 
Besides, SRNet has around one fifth parameters 3.61M of \cite{pavllo20193d}(16.95M) with 243 frame poses as input.

\begin{table*}
\begin{center}
\resizebox{\textwidth}{17mm}{
\begin{tabular}{ l c c c c c c c c c c c c c c c c }
\hline
Method & Direct & Discuss & Eat & Greet & Phone& Photo & Pose  & Purcha. & Sit & SitD &Smoke &Wait &WalkD&Walk& WalkT & Avg.\\
\hline
Hossain et al. \cite{rayat2018exploiting}&48.4& 50.77 &57.2 &55.2& 63.1& 72.6 &53.0& 51.7& 66.1& 80.9& 59.0& 57.3&62.4 &46.6& 49.6& 58.3\\
Lee et al. \cite{lee2018propagating}&\textbf{40.2}& 49.2& 47.8& 52.6& 50.1& 75.0& 50.2& 43.0& 55.8 &73.9 &54.1& 55.6& 58.2& 43.3 &43.3& 52.8\\
Cai et al. \cite{cai2019exploiting}&44.6 &47.4& 45.6& 48.8 &50.8 &59.0 &47.2 &43.9& 57.9& 61.9& 49.7& 46.6& 51.3& 37.1& 39.4& 48.8\\
Pavllo et al. \cite{pavllo20193d}&45.2& 46.7& 43.3 &45.6& 48.1& 55.1 &44.6& 44.3& 57.3& 65.8& 47.1& 44.0 &49.0& 32.8& 33.9& 46.8\\
Lin et al. \cite{lin2019trajectory} &42.5 &\textbf{44.8} &\textbf{42.6}& 44.2& 48.5& 57.1 &\textbf{42.6} &41.4 &56.5 &64.5 &47.4& 43.0& 48.1 &33.0& 35.1& 46.6\\
\emph{Ours} &46.6&47.1&43.9&\textbf{41.6}&\textbf{45.8}&\textbf{49.6}&46.5&\textbf{40.0}&\textbf{53.4}&\textbf{61.1}&\textbf{46.1}&\textbf{42.6}&\textbf{43.1}&\textbf{31.5}&\textbf{32.6}&\textbf{44.8}\\
\hline\hline
Hossain et al. \cite{rayat2018exploiting} $\bigtriangledown$&35.2 &40.8 &37.2 &37.4 &43.2 &44.0 &38.9& 35.6& 42.3& 44.6& 39.7&39.7& 40.2& 32.8& 35.5& 39.2\\
Lee et al. \cite{lee2018propagating} $\bigtriangledown$&32.1 &36.6& 34.3& 37.8& 44.5& 49.9& 40.9& 36.2& 44.1& 45.6& 35.3& 35.9& 37.6& 30.3& 35.5& 38.4\\
Pavllo et al. \cite{pavllo20193d} $\bigtriangledown$ &-&-&-&-&-&-&-&-&-&-&-&-&-&-&-&37.2\\
\emph{Ours} $\bigtriangledown$ &\textbf{34.8} & \textbf{32.1} & \textbf{28.5} &\textbf{30.7} & \textbf{31.4} & \textbf{36.9} & \textbf{35.6} & \textbf{30.5} &\textbf{38.9} &\textbf{40.5} &\textbf{32.5} &\textbf{31.0} & \textbf{29.9}&\textbf{22.5}&\textbf{24.5}&\textbf{32.0}\\
\hline
\end{tabular}}
\end{center}
\caption{Comparison on \emph{Temporal Pose input} in terms of mean per-joint position error (MPJPE). Below the double line, $\bigtriangledown$ indicates use of 2D ground truth pose as input, which is examined to explore the upper bound of these methods. Best results in bold. }
\label{tab:human_p1_t}
\end{table*}

\section{Conclusion}
\label{sec:conclusion}
In this paper, we proposed SRNet, a split-and-recombine approach that improves generalization performance in 3D human pose estimation. The key idea is to design a network structure that splits the human body into local groups of joints and recombines a low-dimensional global context for more effective learning. Experimental results show that SRNet outperforms state-of-the-art techniques, especially for rare and unseen poses. 


\clearpage
%
%
\bibliographystyle{splncs04}
\bibliography{egbib}

\pagebreak
\begin{center}
\textbf{\large ---Supplementary Materials---\\ SRNet: Improving Generalization in 3D Human Pose Estimation with a Split-and-Recombine Approach}
\end{center}
\setcounter{equation}{0}
\setcounter{figure}{0}
\setcounter{table}{0}
\setcounter{section}{0}
\setcounter{page}{0}

In this supplementary material, we present more implementation details of our model, additional experimental results and more qualitative results which are not shown in the main paper due to the space limitation. First, Section \ref{sec:imple} gives more details on experiment settings. Second, Section \ref{sec:recomb} discuses different design choices for the combination operator $\circ$ in Equation 5 of the main paper. Third, Section \ref{sec:p2} shows results of using both MPJPE and PA-MPJPE
as the evaluation metrics in a comparison with state-of-the-art methods on the Human3.6M dataset. Next, Section \ref{sec:cross_action} shows more results under the cross action protocol. Finally, Section \ref{sec:visual} demonstrates additional qualitative results.

\section{Implementation Details} 
\label{sec:imple}

\paragraph{\textbf{Training Data.}} Following many previous works \cite{pavllo20193d,martinez2017simple,fang2018learning,wang2019generalizing,zhao2019semantic,ci2019optimizing}, we show results of using two different kinds of 2D keypoints as input for our model in the experiments. They are 2D ground truth and 2D detections from an off-the-shelf 2D keypoint detector. Following those works \cite{pavllo20193d,cai2019exploiting,lin2019trajectory}, we use the smoothed CPN model \cite{chen2018cascaded} which finetuned on the Human3.6M dataset by an eight-layer residual fully-connected temporal model as our 2D keypoint detector, which is pretrained on the COCO dataset. No extra 2D data has been used for mixed training. In the ablation study (Section 5.1) of the main paper, we use 2D ground truth as input. When comparing with previous works in Section 5.2, both inputs are used and compared respectively. For data normalization, we use two methods. One is provided by \cite{pavllo20193d} called the \emph{\textbf{Basic}} normalization and another is provided by \cite{ci2019optimizing} called the \emph{\textbf{Pixel}} normalization. Please refer to their code base~\cite{tempconvcode,lcncode} for detailed implementation. By default, the \emph{\textbf{Basic}} normalization is used in our ablation study. When comparing with the state-of-the-arts (\cite{ci2019optimizing} in \emph{single pose} and \cite{pavllo20193d} in \emph{temporal pose}), we use the same data normalization method as each for a fair comparison.
For data augmentation, we follow \cite{pavllo20193d,ci2019optimizing} by using horizontal flip data augmentation at both training and test stages.

\paragraph{\textbf{Training Setting.}} Amsgrad \cite{reddi2019convergence} is used as the optimizer. The initial learning rate is 0.001 and it decays by 5\% after each epoch of training. 80 epochs are used in total.
The total channel dimension of each connected/convolution layer is 1024. Batch Normalization~\cite{ioffe2015batch} and Leaky ReLU~\cite{xu2015empirical} activation are applied to each connected/convolution layer. The final network consists of 8 stacked layers, and every two layers (except for the first and last ones) are wrapped with a residual connection as in~\cite{martinez2017simple,pavllo20193d}. Batch size is 1024. L1 loss is used for training.

\section{Design Choices for the Combination Operator}
\label{sec:recomb}
In Equation 5 of the main paper, we show how the low-dimensional global contexts can be brought back to the local group using a combination operator $\circ$. By default, the combination operator $\circ$ is implemented using multiplication in the main paper. Here, we empirically evaluate the design choices of using addition, multiplication and concatenation in Table~\ref{tab:recom}. It is shown that both addition and multiplication obtain favorable results. They surpass the $FC$ and $SFS$ baselines, indicating their effectiveness in recombining the low-dimensional global contexts. 



\begin{table*}
\begin{center}
{
\begin{tabular}{ l| c |c ||c |c |c}
\hline
Method & FC& SFS &SR (add.)&SR (mult.) & SR (concat.)\\
\hline
MPJPE(mm)& 46.8&39.4&36.4&36.6&38.3\\
\hline
Params.(M)&6.39&3.04&1.33&0.88&1.34\\
\hline
\end{tabular}}
\end{center}
\caption{Comparison on different design choices for the combination operator under the
\emph{Subject} protocol. MPJPE is used as the evaluation metric. 2D ground truth is used as input. The third row shows the number of learnable parameters of different models.
}
\label{tab:recom}
\end{table*}

\section{More Results on Human3.6M}
\label{sec:p2}
In Tables 8, 9, and 10 of the main paper, we compare our model with previous works under different input settings (using 2D ground truth or detection, with or without temporal information). We summarise them in Table~\ref{tab:human_p1_single} and \ref{tab:human_p1_temporal} with more detailed results on different actions. Our approach achieves the new state-of-the-art with either 2D keypoint detection or 2D ground truth as input. Specifically, we improve upon \cite{ci2019optimizing} from 36.3mm to 33.9mm (relative 6.6\% improvement) with 2D ground truth input for single pose inputs. We improve upon \cite{lin2019trajectory} from 46.6mm to 44.8mm (relative 3.9\% improvement) with 2D temporal keypoint detection input.

\begin{table*}
\begin{center}
\resizebox{\textwidth}{30mm}
{
\begin{tabular}{ l c c c c c c c c c c c c c c c c }
\hline
Method & Direct & Discuss & Eat & Greet & Phone& Photo & Pose  & Purcha. & Sit & SitD &Smoke &Wait &WalkD&Walk& WalkT & Avg.\\
\hline
Luvizon et al. \cite{luvizon20182d}& 63.8 & 64.0 & 56.9 & 64.8 & 62.1 & 70.4 & 59.8 & 60.1&71.6 & 91.7 &60.9 & 65.1 & 51.3 & 63.2 & 55.4 & 64.1  \\ 
Martinez et al. \cite{martinez2017simple}& 51.8 &56.2 &58.1 &59.0 &69.5 &78.4 &55.2 &58.1& 74.0& 94.6 &62.3 &59.1& 65.1& 49.5& 52.4 &62.9 \\
Park et al.\cite{park20183d}&49.4 &54.3& 51.6& 55.0& 61.0 &73.3 &53.7 &50.0& 68.5& 88.7 &58.6 &56.8& 57.8& 46.2 &48.6 &58.6\\
Wang et al. \cite{wang2019generalizing}&47.4 &56.4 &49.4 &55.7& 58.0& 67.3& \textbf{46.0}& 46.0& 67.7& 102.4& 57.0& 57.3& 41.1& 61.4& 40.7 &58.0\\
Zhao et al. \cite{zhao2019semantic}&47.3 &60.7& 51.4& 60.5& 61.1 &\textbf{49.9} &47.3 &68.1& 86.2& \textbf{55.0}& 67.8& 61.0& 42.1& 60.6& 45.3& 57.6\\
Ci et al. \cite{ci2019optimizing}&46.8 &52.3& 44.7& 50.4& 52.9& 68.9& 49.6& 46.4& 60.2 &78.9& \textbf{51.2}& 50.0& 54.8& 40.4& 43.3& 52.7\\
Pavllo et al. \cite{pavllo20193d} &47.1& 50.6& 49.0& 51.8 &53.6 &61.4& 49.4 &47.4 &59.3 &67.4 &52.4& 49.5& 55.3& 39.5& 42.7& 51.8\\
Cai et al. \cite{cai2019exploiting}&46.5 &48.8 &47.6& 50.9& 52.9 &61.3 &48.3 &45.8 &\textbf{59.2} &64.4& \textbf{51.2}& 48.4& \textbf{53.5}& 39.2& 41.2& 50.6\\
\emph{Ours} &\textbf{44.5}&\textbf{48.2}&\textbf{47.1}&\textbf{47.8}&\textbf{51.2}&56.8&50.1&\textbf{45.6}&59.9&66.4&52.1&\textbf{45.3}&54.2&\textbf{39.1}&\textbf{40.3}&\textbf{49.9}\\
\hline
\hline
Martinez et al. \cite{martinez2017simple}$\bigtriangledown$ &37.7 &44.4 &40.3 &42.1& 48.2& 54.9 &44.4 &42.1 &54.6& 58.0 &45.1 &46.4 &47.6 &36.4 &40.4& 45.5\\
Pham et al. \cite{pham2019unified}$\bigtriangledown$ &36.6 &43.2 &38.1 &40.8 &44.4 &51.8 &43.7 &38.4 &50.8 &52.0 &42.1 &42.2 &44.0 &32.3 &35.9 &42.4\\
Zhao et al. \cite{zhao2019semantic}$\bigtriangledown$ &37.8 &49.4 &37.6 &40.9 &45.1 &\textbf{41.4} &40.1& 48.3& 50.1 &42.2& 53.5 &44.3 &40.5 &47.3& 39.0 &43.8\\
Wang et al. \cite{wang2019generalizing}$\bigtriangledown$& 35.6 &41.3 &39.4 &40.0 &44.2 &51.7 &39.8 &40.2 &50.9 &55.4 &43.1 &42.9 &45.1 &33.1 &37.8 &42.0\\
\emph{Ours}-Basic $\bigtriangledown$ &35.9&36.7&29.3&34.5&36.0&42.8&37.7&31.7&40.1&44.3&35.8&37.2&36.2&33.7&34.0&36.4\\
Ci et al.-Pixel \cite{ci2019optimizing} $\bigtriangledown$&36.3& 38.8 &29.7& 37.8& 34.6& 42.5& 39.8& 32.5& \textbf{36.2}& \textbf{39.5}& 34.4& 38.4& 38.2& 31.3& 34.2&36.3\\
\emph{Ours}-Pixel $\bigtriangledown$&\textbf{32.9}&\textbf{34.5}&\textbf{27.6}&\textbf{31.7}&\textbf{33.5}&42.5&\textbf{35.1}&\textbf{29.5}&38.9&45.9&\textbf{33.3}&\textbf{34.9}&\textbf{34.4}&\textbf{26.5}&\textbf{27.1}&\textbf{33.9}\\
\hline
\end{tabular}}
\end{center}
\caption{Detailed \emph{single pose} comparison in terms of the mean per-joint position error (MPJPE) on Human3.6M. Below the double line are results from 2d ground truth inputs (indicated by $\bigtriangledown$) to explore the upper bound of these methods. Best results in bold. }
\label{tab:human_p1_single}
\end{table*}

\begin{table*}
\begin{center}
\resizebox{\textwidth}{20mm}
{
\begin{tabular}{ l c c c c c c c c c c c c c c c c }
\hline
Method & Direct & Discuss & Eat & Greet & Phone& Photo & Pose  & Purcha. & Sit & SitD &Smoke &Wait &WalkD&Walk& WalkT & Avg.\\
\hline
Hossain et al. \cite{rayat2018exploiting} &48.4& 50.77 &57.2 &55.2& 63.1& 72.6 &53.0& 51.7& 66.1& 80.9& 59.0& 57.3&62.4 &46.6& 49.6& 58.3\\
Lee et al. \cite{lee2018propagating} &\textbf{40.2}& 49.2& 47.8& 52.6& 50.1& 75.0& 50.2& 43.0& 55.8 &73.9 &54.1& 55.6& 58.2& 43.3 &43.3& 52.8\\
Cai et al. \cite{cai2019exploiting} &44.6 &47.4& 45.6& 48.8 &50.8 &59.0 &47.2 &43.9& 57.9& 61.9& 49.7& 46.6& 51.3& 37.1& 39.4& 48.8\\
Pavllo et al. \cite{pavllo20193d}&45.2& 46.7& 43.3 &45.6& 48.1& 55.1 &44.6& 44.3& 57.3& 65.8& 47.1& 44.0 &49.0& 32.8& 33.9& 46.8\\
Lin et al. \cite{lin2019trajectory} &42.5 &\textbf{44.8} &\textbf{42.6}& 44.2& 48.5& 57.1 &\textbf{42.6} &41.4 &56.5 &64.5 &47.4& 43.0& 48.1 &33.0& 35.1& 46.6\\
\emph{Ours} &43.1&47.1&43.9&\textbf{41.6}&\textbf{45.8}&\textbf{49.6}&46.5&\textbf{40.0}&\textbf{53.4}&\textbf{61.1}&\textbf{46.1}&\textbf{42.6}&\textbf{46.6}&\textbf{31.5}&\textbf{32.6}&\textbf{44.8}\\
\hline\hline
Hossain et al. \cite{rayat2018exploiting} $\bigtriangledown$&35.2 &40.8 &37.2 &37.4 &43.2 &44.0 &38.9& 35.6& 42.3& 44.6& 39.7&39.7& 40.2& 32.8& 35.5& 39.2\\
Lee et al. \cite{lee2018propagating} $\bigtriangledown$ &32.1 &36.6& 34.3& 37.8& 44.5& 49.9& 40.9& 36.2& 44.1& 45.6& 35.3& 35.9& 37.6& 30.3& 35.5& 38.4\\
Pavllo et al.-243f \cite{pavllo20193d} $\bigtriangledown$ &-&-&-&-&-&-&-&-&-&-&-&-&-&-&-&37.2\\

\emph{Ours}-243f $\bigtriangledown$ &\textbf{34.8} & \textbf{32.1} & \textbf{28.5} &\textbf{30.7} & \textbf{31.4} & \textbf{36.9} & \textbf{35.6} & \textbf{30.5} &\textbf{38.9} &\textbf{40.5} &\textbf{32.5} &\textbf{31.0} &\textbf{29.9}&\textbf{22.5}&\textbf{24.5}&\textbf{32.0}\\
\hline
\end{tabular}}
\end{center}
\caption{Detailed \emph{temporal pose} comparison in terms of the mean per-joint position error (MPJPE) on Human3.6M. Below the double line are results from 2d ground truth inputs (indicated by $\bigtriangledown$) to explore the upper bound of these methods. Best results in bold. }
\label{tab:human_p1_temporal}
\end{table*}

In Table \ref{tab:human_p2_single} and \ref{tab:human_p2_temporal}, we compare with the previous works using the \emph{PA-MPJPE} metric where available. Our approach achieves the new state-of-the-art with either 2D keypoint detection or 2D ground truth (denoted by $\bigtriangledown$) as input. Specifically, we improve upon \cite{ci2019optimizing} from 27.9mm to 24.3mm (relative 14.8\% improvement) with 2D ground truth input. We improve upon \cite{pavllo20193d} from 36.5mm to 34.9mm (relative 4.4\% improvement) with 2D keypoint detection input.


\begin{table*}
\begin{center}
\resizebox{\textwidth}{16mm}{
\begin{tabular}{ l c c c c c c c c c c c c c c c c }
\hline
Method & Direct & Discuss & Eat & Greet & Phone& Photo & Pose  & Purcha. & Sit & SitD &Smoke &Wait &WalkD&Walk& WalkT & Avg.\\
\hline
Martinez et al. \cite{martinez2017simple} & 39.5 &43.2& 46.4 &47.0 &51.0 &56.0 &41.4 &40.6& 56.5& 69.4 &49.2 &45.0 &49.5 &38.0 &43.1 &47.7  \\

Fang et al.\cite{fang2018learning}&38.2 &41.7 &43.7 &44.9 &48.5 &55.3 &40.2 &38.2 &54.5 &64.4 &47.2& 44.3 &47.3 &36.7& 41.7& 45.7\\
Park et al. \cite{park20183d}&38.3 &42.5& 41.5& 43.3 &47.5& 53.0& 39.3 &37.1& 54.1& 64.3& 46.0& 42.0& 44.8& 34.7& 38.7& 45.0\\

Ci et al. \cite{ci2019optimizing} &36.9& 41.6 &38.0 &41.0 &41.9 &51.1& 38.2& 37.6& 49.1 &62.1 &43.1& 39.9& 43.5& 32.2& 37.0& 42.2\\
Pavlakos et al. \cite{pavlakos2018ordinal}&\textbf{34.7}& 39.8& 41.8& 38.6 &42.5 &47.5 &38.0& 36.6& 50.7& 56.8& 42.6& 39.6& 43.9& 32.1& 36.5& 41.8\\
Pavllo et al.\cite{pavllo20193d}&36.0 &38.7& 38.0& 41.7& 40.1 &45.9 &\textbf{37.1} &\textbf{35.4} &\textbf{46.8} &\textbf{53.4}& 41.4& 36.9 &43.1 &30.3 &\textbf{34.8} &40.0\\

Ours&35.8&\textbf{39.2}&\textbf{36.6}&\textbf{36.9}&\textbf{39.8}&\textbf{45.1}&38.4&36.9&47.7&54.4&\textbf{38.6}&\textbf{36.3}&\textbf{39.4}&\textbf{30.3}&35.4&\textbf{39.4}\\

\hline\hline
\emph{Ours}-Basic $\bigtriangledown$
&26.0&28.9&23.7&26.9&27.4&33.1&27.9&25.0&32.4&40.9&28.8&29.2&29.3&23.3&24.5&28.5\\
\emph{Ours}-Pixel $\bigtriangledown$ &24.1&28.6&24.2&26.6&26.3&35.1&27.7&24.5&32.8&39.1&27.8&28.0&29.6&22.3&23.0&28.0\\
\hline

\hline
\end{tabular}}
\end{center}
\caption{Comparison \emph{single pose} results regarding PA-MPJPE after rigid transformation from the ground truth.$\bigtriangledown$ indicates the use of 2D ground truth poses as input. Best results in bold.}
\label{tab:human_p2_single}
\end{table*}

\begin{table*}
\begin{center}
\resizebox{\textwidth}{14mm}{
\begin{tabular}{ l c c c c c c c c c c c c c c c c }
\hline
Method & Direct & Discuss & Eat & Greet & Phone& Photo & Pose  & Purcha. & Sit & SitD &Smoke &Wait &WalkD&Walk& WalkT & Avg.\\
\hline
Lee et al.\cite{lee2018propagating}&38.0 &39.3 &46.3 &44.4 &49.0 &55.1 &40.2& 41.1& 53.2 &68.9 &51.0 &39.1 &\textbf{33.9} &56.4 &38.5 &46.2\\
Hossain et al.\cite{rayat2018exploiting}& 35.7 &39.3 &44.6 &43.0 &47.2 &54.0 &38.3& 37.5& 51.6 &61.3 &46.5 &41.4& 47.3 &34.2 &39.4& 44.1\\
Cai et al.\cite{cai2019exploiting}&35.7 &37.8 &36.9& 40.7& 39.6 &45.2 &37.4 &34.5 &46.9 &\textbf{50.1}& 40.5 &36.1 &41.0 &29.6& 33.2& 39.0\\
Lin et al. \cite{lin2019trajectory}&32.5& 35.3& \textbf{34.3}& 36.2& 37.8& 43.0& \textbf{33.0}& 32.2 &45.7 &51.8& 38.4& 32.8& 37.5& 25.8 &28.9& 36.8\\
Pavllo et al.-243f \cite{pavllo20193d}&34.1 &36.1& 34.4& 37.2&36.4& \textbf{42.2} &34.4 &33.6&45.0& 52.5& 37.4& 33.8& 37.8& \textbf{25.6}& 27.3& 36.5\\
\emph{Ours}-243f&\textbf{31.9}&\textbf{33.7}&34.7&\textbf{35.0}&\textbf{35.5}&42.8&36.4&\textbf{30.5}&\textbf{43.6}&51.3&\textbf{36.7}&\textbf{32.5}&\textbf{36.5}&27.5&\textbf{25.7}&\textbf{34.9}\\
\hline
\hline
\emph{Ours}-243f $\bigtriangledown$ &\textbf{23.7}&\textbf{25.2}&\textbf{22.9}&\textbf{23.1}&\textbf{24.0}&\textbf{28.7}&\textbf{25.0}&\textbf{22.1}&\textbf{31.8}&\textbf{32.8}&\textbf{24.8}&\textbf{23.5}&\textbf{23.4}&\textbf{17.0}&\textbf{18.3}&\textbf{24.3}\\
\hline
\end{tabular}}
\end{center}
\caption{Comparison \emph{temporal pose} results regarding PA-MPJPE after rigid transformation from the ground truth. $243f$ means inputs contain 243 frame poses. $\bigtriangledown$ indicates the use of 2D ground truth poses as input. Best results in bold.}
\label{tab:human_p2_temporal}
\end{table*}

\section{Cross Action Results Using 2D Ground Truth Input}
\label{sec:cross_action}
In Table 7 of the main paper, we compare our cross-action results with \cite{ci2019optimizing} under the same data settings.
Here, we provide more results of using 2d ground truth as input under the cross-action protocol. The FCN baseline~\cite{martinez2017simple} (with our implementation) and our SRNet are compared. Both MPJPE and PA-MPJPE (with $\times$) are used as the evaluation metrics. Both \emph{\textbf{Basic}} and \emph{\textbf{Pixel}} \cite{ci2019optimizing} normalization results of our method are reported.

In Table~\ref{tab:ca_gt}, our method gains improvements in terms of MPJPE from 80.6mm to 64.3mm, by 16.3mm (relatively 20.2\%). For PA-MPJPE, the improvement is from 60.5mm to 49.4mm, by 11.1mm (relatively 18.3\%).

\begin{table*}
\begin{center}
\resizebox{\textwidth}{10mm}{
\begin{tabular}{ l c c c c c c c c c c c c c c c c }
\hline
Method & Direct & Discuss & Eat & Greet & Phone& Photo & Pose  & Purcha. & Sit & SitD &Smoke &Wait &WalkD&Walk& WalkT & Avg.\\
\hline
FCN-Pixel \cite{martinez2017simple} &117.0&67.4&62.6&93.0&59.5&72.8&66.7&80.0&71.2&71.6&58.6&75.2&73.3&114.9&125.0&80.6\\
Ours-Basic &91.1&54.8&59.0&71.2&50.9&61.5&65.0&71.4&76.6&74.0&50.3&64.8&58.1&78.0&85.8&67.5\\
Ours-Pixel&\textbf{86.2}&\textbf{53.0}&\textbf{55.0}&\textbf{70.5}&\textbf{47.9}&\textbf{57.9}&\textbf{63.1}&\textbf{68.4}&\textbf{71.2}&\textbf{72.9}&\textbf{47.5}&\textbf{59.4}&\textbf{56.3}&\textbf{70.8}&\textbf{83.8}&\textbf{64.3}\\
\hline
FCN-Pixel\cite{martinez2017simple} $\times$&91.9&55.3&51.8&75.2&49.3&60.6&57.3&64.7&62.2&60.6&49.5&62.7&61.3&95.4&99.8&60.5\\
Ours-Basic $\times$&65.9&42.4&46.3&54.5&39.8&46.6&50.6&55.8&58.4&57.4&39.3&49.6&45.0&56.7&61.8&51.3\\
Ours-Pixel$\times$&\textbf{61.7}&\textbf{42.0}&\textbf{44.2}&\textbf{53.1}&\textbf{38.5}&\textbf{45.2}&\textbf{49.5}&\textbf{53.6}&\textbf{55.5}&\textbf{55.5}&\textbf{37.9}&\textbf{46.4}&\textbf{43.8}&\textbf{54.7}&\textbf{59.7}&\textbf{49.4}\\
\hline
\end{tabular}}
\end{center}
\caption{\emph{Cross Action} comparison to the FCN baseline with \emph{2D ground truth input} on Human3.6M in terms of mean per-joint position error (MPJPE) and PA-MPJPE (denoted by $\times$).}
\label{tab:ca_gt}
\end{table*}

\section{Additional Qualitative Results}
\label{sec:visual}
\begin{figure}
\begin{center}
\includegraphics[width=1.0\textwidth,trim=40 210 400 160,clip]{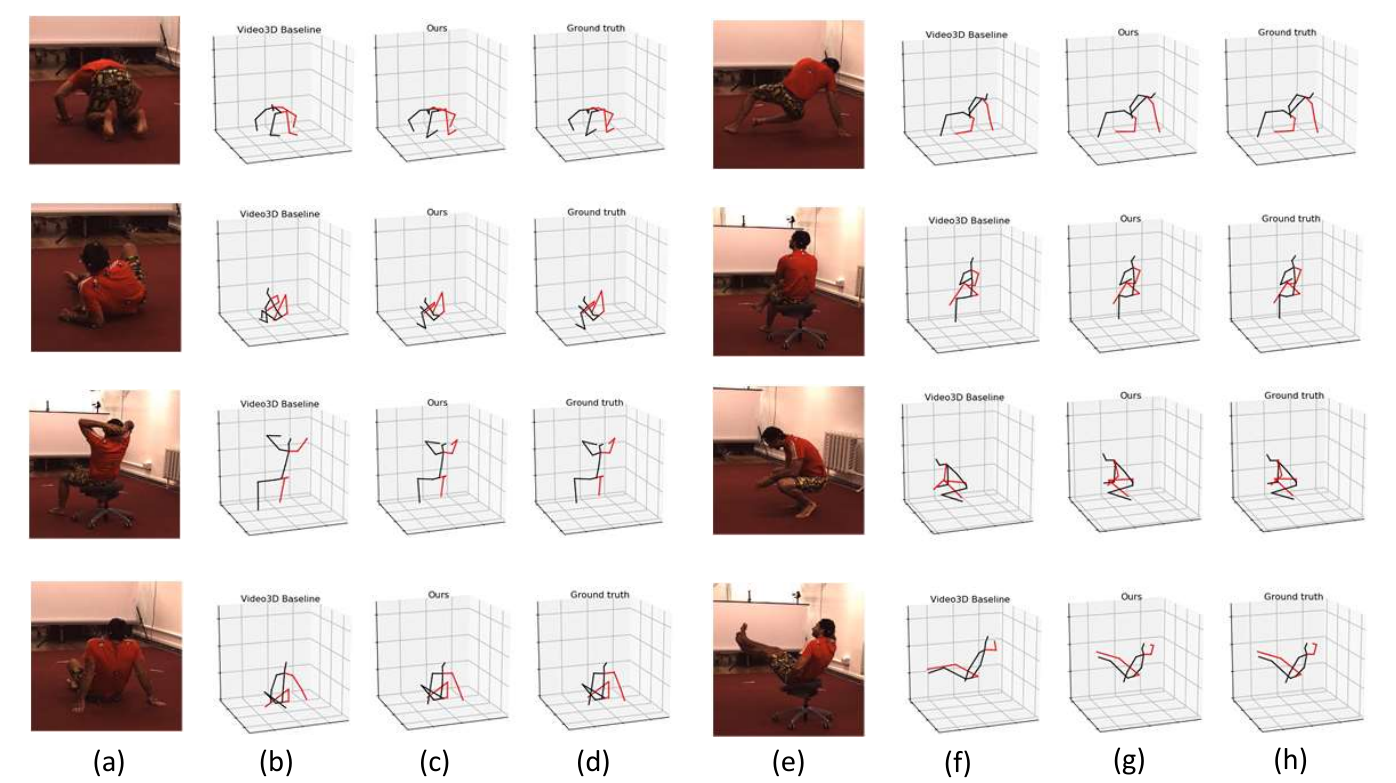}
\end{center}
\caption{Visualization results trained with the subject protocol settings on the Human3.6M dataset. (a), (e) are the original test images. (b), (f) show the 3D pose predictions of temporal 3D pose baseline \cite{pavllo20193d}. (c), (g) are the 3D pose predictions of our method. (d), (h) are the 3D ground truth poses.}
\label{fig:vis_hm}
\end{figure}

\begin{figure}
\begin{center}
\includegraphics[width=1.0\textwidth,trim=0 170 270 70,clip]{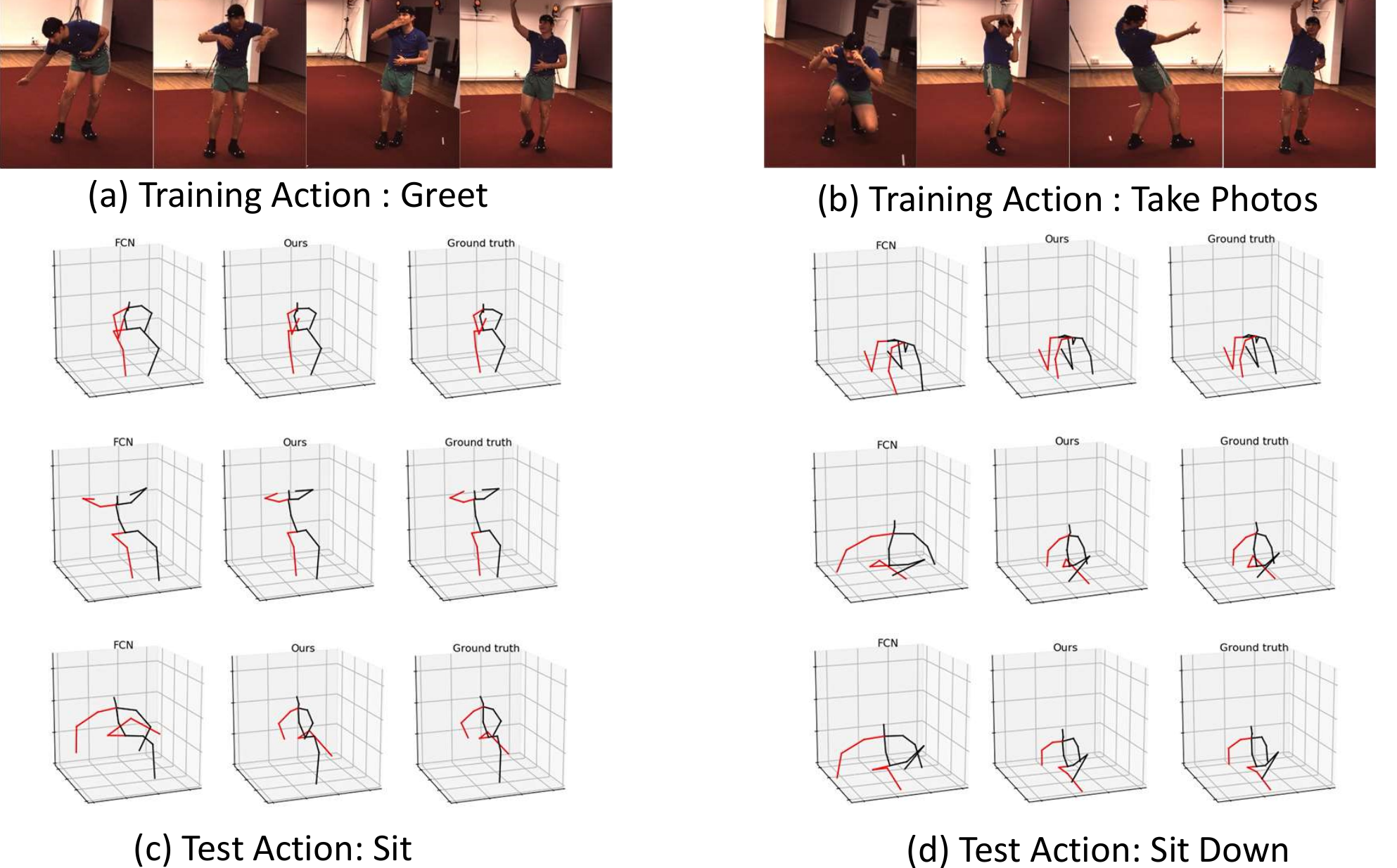}
\end{center}
\caption{Visualization results for the cross-action protocol. (a), (b) are two kinds of original training actions. (c), (d) show the 3D predicted results by FCN \cite{martinez2017simple}, our method, and the 3D ground truth poses on two kinds of test actions. When training action is “greet”, poses like in (a), test on the action “sit” to get those predictions in (c). Similarly, when training action is “take photos” in (b), test on the action “sit down” to show the differences between the FCN and our method in (d). }
\label{fig:vis_ca}
\end{figure}

\begin{figure}
\begin{center}
\includegraphics[width=1.0\textwidth,trim=0 350 450 60,clip]{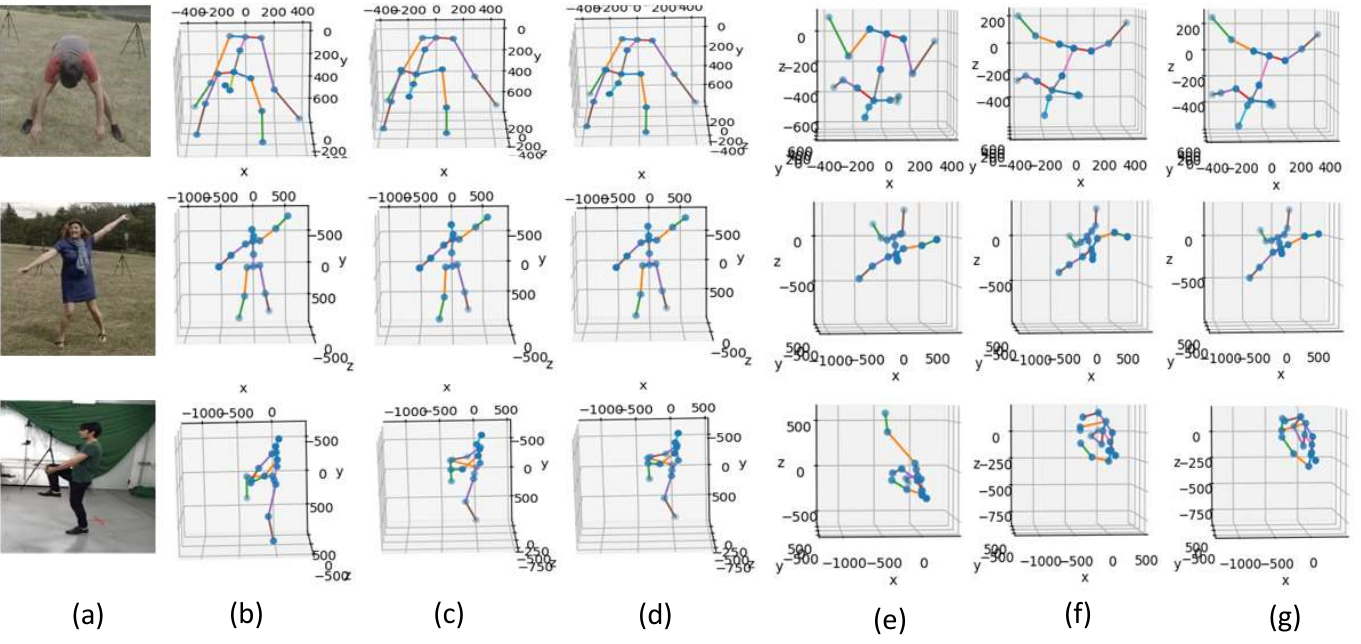}
\end{center}
\caption{Visualization results for the MPI-INF-3DHP dataset. (a) are the original images. (b), (e) show the 3D predicted results by \cite{martinez2017simple} from the front viewpoint and the top viewpoint. (c), (f) show the prediction poses of our method, and (d), (g) are the 3D ground truth poses from the front viewpoint and the top viewpoint, separately.}
\label{fig:vis_mpi}
\end{figure}

Besides the aforementioned quantitative results, we also present some qualitative results. First, we visualize some hard poses, which are also rare in the subject protocol evaluation, in Figure \ref{fig:vis_hm}. Under this protocol, our method can predict well even on challenging poses such as kowtow, side-lying and legs lifting. Next, Figure \ref{fig:vis_ca} demonstrates some unseen poses in the cross-action protocol to verify our method's generalization ability. 
Finally, Figure \ref{fig:vis_mpi} shows some qualitative results with training only on the Human3.6M dataset and testing on unseen poses and unseen camera angles. Nevertheless, our method is still able to reconstruct many plausible 3D poses well.

\clearpage

\end{document}